\documentclass[11pt]{article}

\usepackage[preprint]{main}

\usepackage{times}
\usepackage{latexsym}
\usepackage{booktabs}
\usepackage{amsmath}
\usepackage{tabularx}
\usepackage{subcaption}
\usepackage{float}
\usepackage{placeins}
\newcommand{\valp}[2]{\shortstack[c]{#1\\(#2)}}

\usepackage[T1]{fontenc}

\usepackage[utf8]{inputenc}

\usepackage{microtype}

\usepackage{inconsolata}

\usepackage{graphicx}

\title{What if I ask in \textit{alia lingua}? \\Measuring Functional Similarity Across Languages}

\author{
Debangan Mishra\textsuperscript{*1} \quad
Arihant Rastogi\textsuperscript{*1} \quad
Agyeya Negi\textsuperscript{1} \\
\bf{Shashwat Goel\textsuperscript{2}} \quad
\bf{Ponnurangam Kumaraguru\textsuperscript{1}} \\
\textsuperscript{1}IIIT Hyderabad \quad
\textsuperscript{2}ELLIS Institute Tübingen
}

\begin{document}
\maketitle
\begin{abstract}
How similar are model outputs across languages? In this work, we study this question using a recently proposed model similarity metric $\kappa_p$ applied to 20 languages and 47 subjects in GlobalMMLU. Our analysis reveals that a model's responses become increasingly consistent across languages as its size and capability grow. Interestingly, models exhibit greater cross-lingual consistency within themselves than agreement with other models prompted in the same language. These results highlight not only the value of $\kappa_p$ as a practical tool for evaluating multilingual reliability, but also its potential to guide the development of more consistent multilingual systems.

\end{abstract}

\section{Introduction}

Users interact with large language models (LLMs) in a variety of languages across families and resource availabilities ~\citep{nicholas2023lost}. As such, there is a need for LLMs to perform well across languages. These models should provide consistent responses—if switching languages results in incorrect answers to the same question, it could potentially mislead users, especially in critical areas like medical advice or legal interpretation. However, current evaluations primarily focus on per-language accuracy, with little attention to consistency across languages ~\citep{koto2024arabicmmlu, romanou2024include, singh2024global}.

To quantify this consistency, we study the functional similarity of model outputs. We use Chance Adjusted Probabilistic Agreement (CAPA or $\kappa_p$), a metric recently proposed by ~\citep{goel2025great}, which incorporates model accuracy on a given benchmark. We extend it to measure how similar the mistakes are across different languages, giving a view of multilingual functional similarity.

\begin{figure}
    \centering
    \includegraphics[width=\columnwidth]{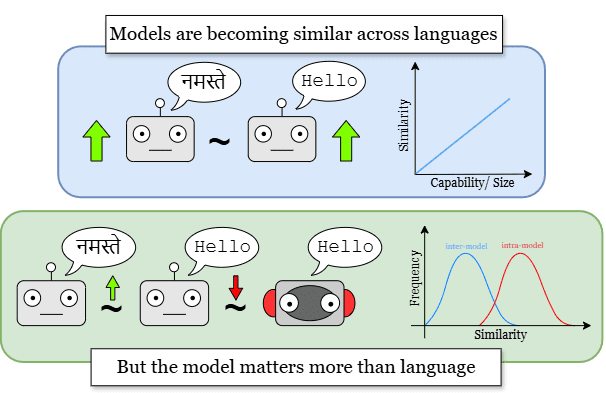}
    \caption{\textbf{Our Main Findings:} We use functional similarity to measure the consistency of model outputs across different languages. We find: (1) as language models get bigger and more capable, their outputs become more similar across languages; (2) models tend to be more self-consistent across languages than when comparing different models in a common language.}
    \label{fig:main-findings}
\end{figure}

\renewcommand\thefootnote{}
\footnote{* \textit{These authors contributed equally}}
\footnote{Please find our code here: \href{https://github.com/Debangan-MishraIIIT/multilingual_evaluations}{GitHub}}
\renewcommand\thefootnote{\arabic{footnote}}

We use GlobalMMLU ~\citep{singh2024global} - a carefully translated version of MMLU across multiple languages -  as our benchmark. It tests the factual QA capabilities of models across a variety of subjects, ranging from mathematics to philosophy, in a multiple-choice format. Our choice of this benchmark is motivated by its parallel nature, which allows us to test whether models behave consistently across languages on factual tasks.

Our study encompasses two dimensions of functional similarity: intra-model (consistency across languages for a given model) and inter-model (consistency across models for a given language). When considering intra-model similarity, we find that with increasing size and accuracy, models are becoming more functionally similar across languages. Notably, we observe that all models are more consistent with themselves across languages than they are with other LLMs for the same language, indicating that intra-model similarity exceeds inter-model similarity for our task. Interestingly, multilingual similarity further varies by domain and resource levels of the languages.
 
Primarily, we show that $\kappa_p$, a chance-adjusted functional similarity metric, provides a powerful lens for analyzing multilingual consistency of LLMs. We explore cross-lingual patterns that accuracy and representational similarity alone cannot capture, by combining the output behavior and performance of the LLM. To illustrate this, we uncover interesting patterns about multilingual model behavior, including effects of scale, domain, and resources.

\section{Related Work}

\textbf{Similarity Metrics:} Prior work on model similarity falls broadly into two classes: \textit{representational similarity} and \textit{functional similarity}. Representational similarity metrics \citep{huh2024platonic, klabunde2025similarity} focus on the internal states of models such as weights and activations, whereas functional similarity metrics \citep{goel2025great} evaluate models based on their input–output behavior, making them applicable across architectures. Importantly, functional similarity better reflects the user experience, since what ultimately matters is whether models behave consistently across inputs, rather than how their internal representations align.

\textbf{Multilingual Evaluations:} In representational studies, researchers have identified language-specific neurons \citep{tang2024language} and language-agnostic “semantic hubs” \citep{semantichub}, and even used steering interventions to demonstrate their causal effects. While such work sheds light on cross-lingual representations, it does not establish quantitative trends in cross-lingual \textit{output} consistency as models scale. On the functional side, prior work on multilingual factual consistency \citep{qi2023cross}, as well as classical agreement metrics \citep{scott1955pi, cohen1960coefficient}, do not account for model accuracy and can overestimate similarity. This leaves a gap for metrics such as $\kappa_p$, which explicitly account for error consistency with agreement to provide a more realistic view of multilinguality.

\section{Methodology}
The accuracy of LLMs differ greatly across languages and their performance is particularly in low-resource languages~\citep{li2025language}. This can artificially inflate similarity scores for some languages as high performance leaves little room for disagreement (as explained further in Appendix \ref{sec:capa_usage}). Given that $\kappa_p$ addresses these issues, we use it to compare similarity of model outputs in light of variable performance across languages. Our work complements studies on representational similarity across languages such as \citet{wu2024semantic}.
 
$\kappa_p$ computes observed agreement $c_{\text{obs}}^p$ as the proportion with which the same option is selected across samples. To account for agreement by chance, $\kappa_p$ introduces an expected agreement $c_{\text{exp}}^p$, derived from the marginal distribution of each set of predictions. The $\kappa_p$ score is given by:
\[
\kappa_p = \frac{c_{\text{obs}}^p - c_{\text{exp}}^p}{1 - c_{\text{exp}}^p},
\]
As $\kappa_p$ increases, models make more similar mistakes, and their errors become more correlated, making them functionally more similar. Henceforth, we compute the average $\kappa_p$ using micro-averaging by concatenating all datasets in the group and then computing the $\kappa_p$ across the combined set. Since $\kappa_p$ is non-linear, the technique of micro-averaging is preferred as it smooths out extremes and operates directly at the per-sample level to better understand $\kappa_p$ across a dataset. 

We use Gemma-3 (1B, 4B and 12B variants) \citep{team2025gemma3} and Qwen-3 (1.7B, 4B, 8B and 14B variants) \citep{yang2025qwen3} in our experiments, as they are some of the latest models as of August 2025 which have undergone multilingual pretraining. We also use the older Gemma-7B \citep{team2024gemma} as a sanity check. We evaluate these models on a subset of 20 languages of the GlobalMMLU dataset~\citep{singh2024global} with our choice of languages justified in Appendix~\ref{sec:lang_choice}. Building on our evaluation methodology, we leverage the LM Evaluation Harness \citep{eval-harness}, a unified framework for testing generative language models on a wide variety of benchmarks known for its reproducibility and extensive adoption.

\section{Experimentation}
\subsection{Intra-Model Multilingual Similarity}
\begin{figure}[H]
    \centering
    \includegraphics[width=\columnwidth]{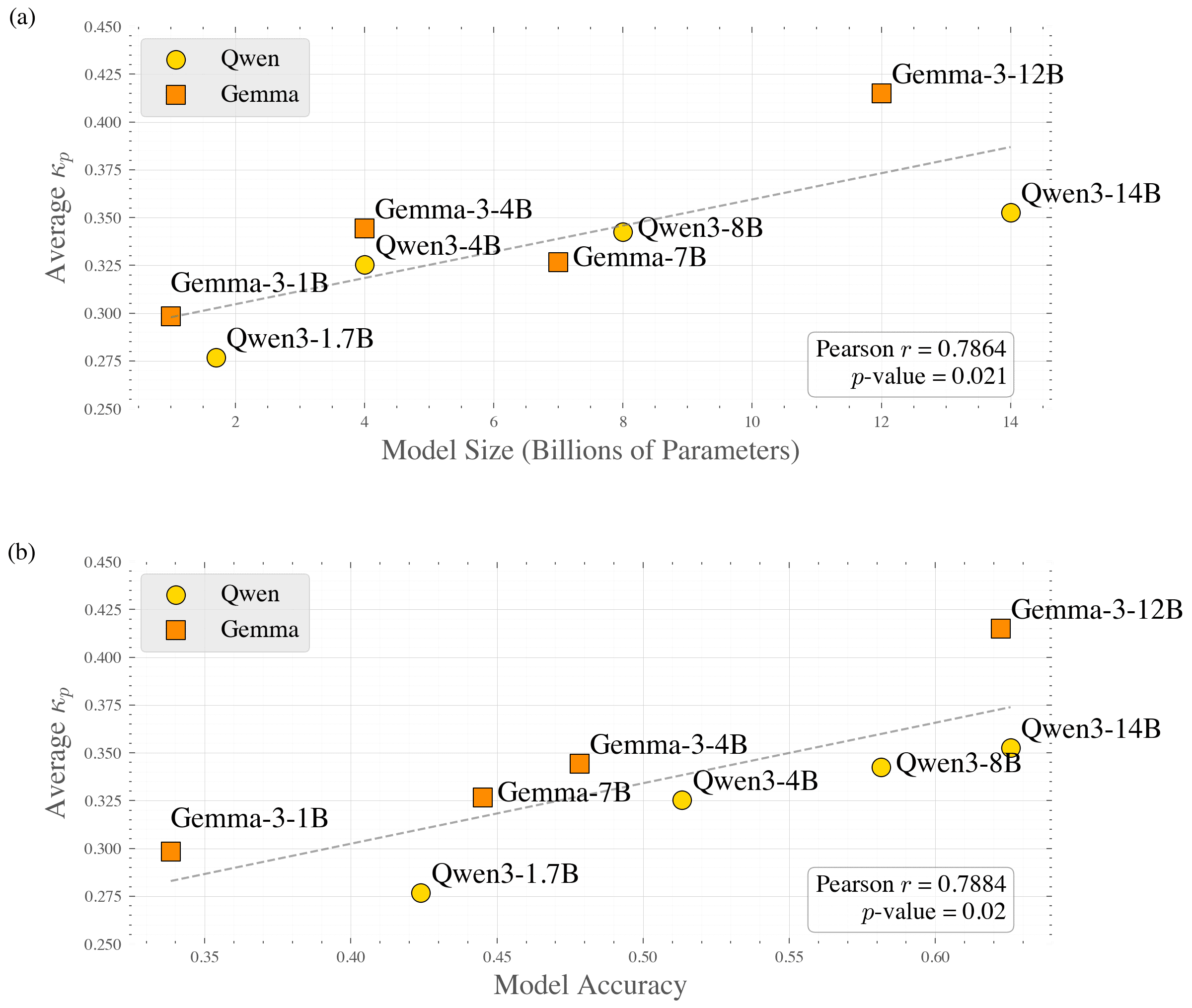}
    \caption{$\kappa_p$ correlates positively with model size and accuracy. (a) $\kappa_p$ averaged over languages positively correlates with model size (b) Similarly, $\kappa_p$ averaged over languages positively correlates with model performance. This indicates that models grow similar across languages with their capability and size.}
    \label{fig:capa_v_acc}
\end{figure}

\textbf{RQ1: Are LLMs becoming similar across languages?} Motivated by the findings of \citet{huh2024platonic} which shows that model representations tend to converge with an increase in size and performance of models, we investigate whether a similar convergence occurs in the output space across languages. A clear trend is observed --- as the model size increases, the average $\kappa_p$ score across languages also increases. $\kappa_p$ also positively correlates with model accuracy. These findings suggest that outputs become more consistent across languages for larger and more accurate LLMs. The statistically significant results are illustrated in Figure~\ref{fig:capa_v_acc}. A possible reason for this could be that bigger models are trained on a greater volume of
data including from low resource languages allowing for greater similarity. But it is not possible to confirm this hypothesis as we do not have access to their exact training data. 

\begin{figure}[H]
    \centering
    \includegraphics[width=\columnwidth]{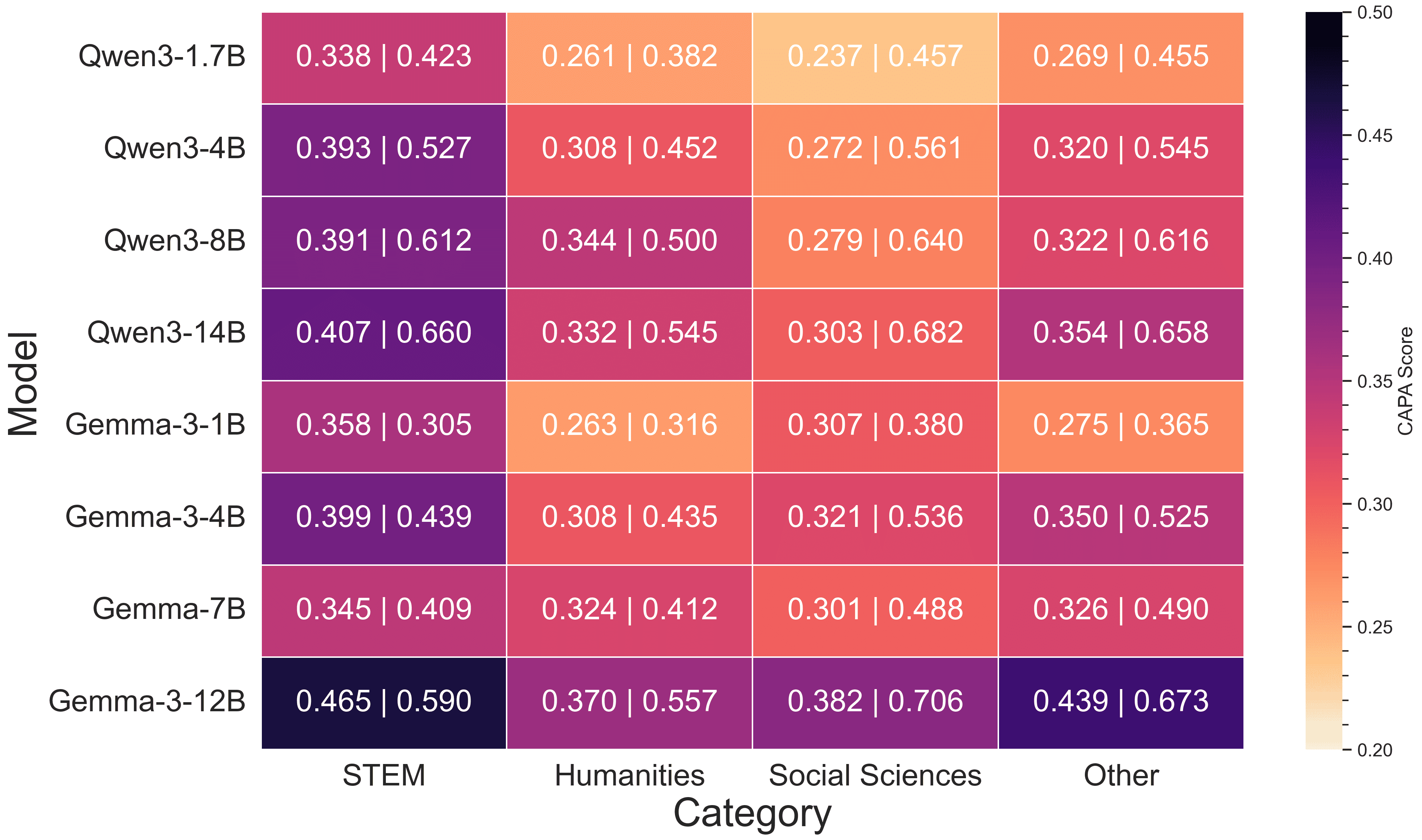}
    \caption{Models answer more similarly across languages for STEM than other domains. Each heatmap cell represents the $\kappa_p$ and accuracy averaged over languages. For example, a cell value of (0.3 | 0.4) for a given model and category would represent an average $\kappa_p$ of 0.3 and an average accuracy of 40\%, both averaged over all the languages.}
    \label{fig:subject_capa}
\end{figure}

\begin{figure}[H]
    \centering
    \includegraphics[width=\columnwidth]{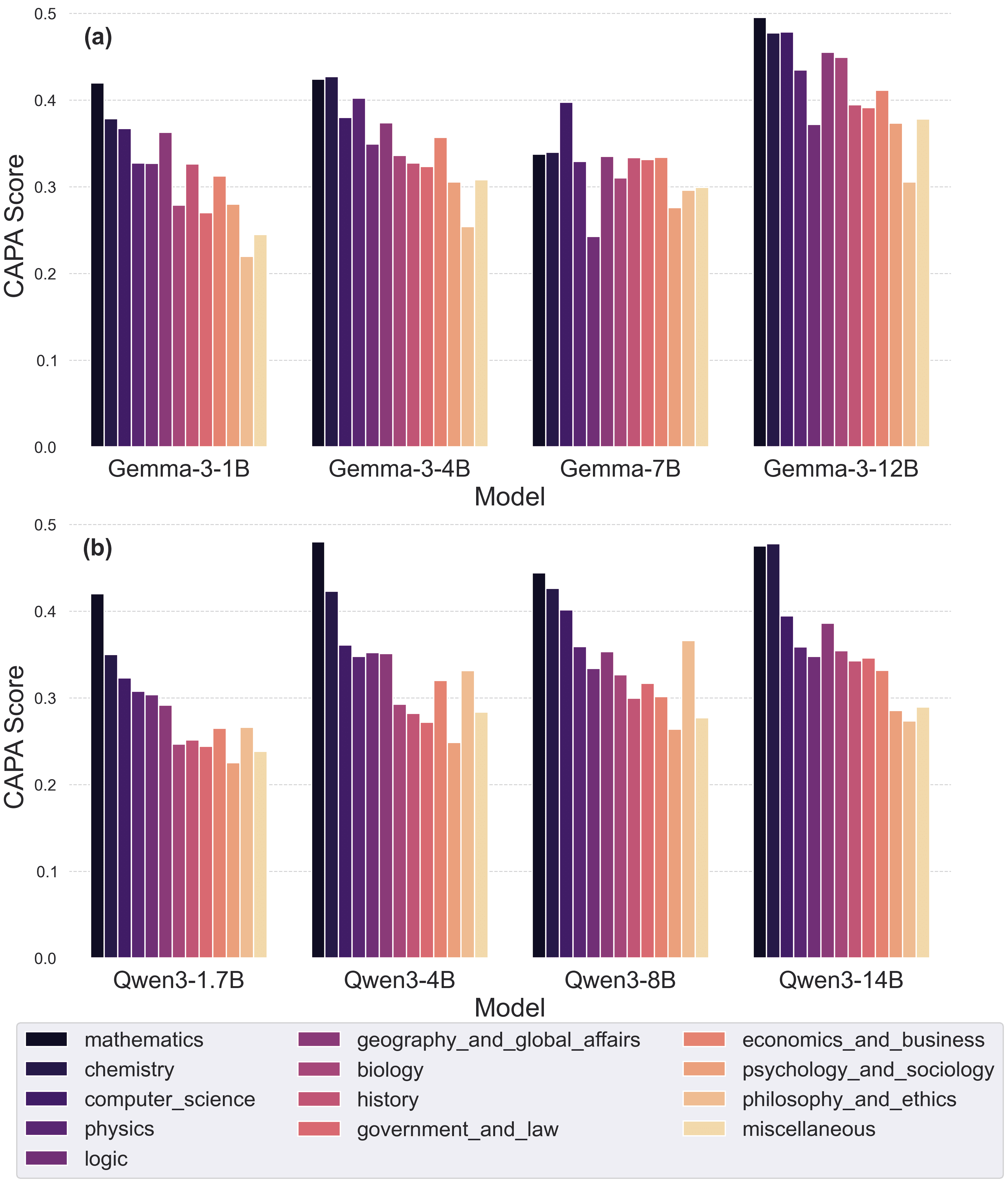}
    \caption{Intra-model $\kappa_p$ scores are higher for categories belonging to STEM (Mathematics, Physics, Computer Science) than the Humanities (Philosophy, Psychology, Sociology). (a) Family of Gemma models (b) Family of Qwen Models.}
    \label{fig:capa_scores_gemma_qwen_vertical}
\end{figure}

\textbf{RQ2: Does the domain of questions asked matter?}  Prior work shows that the language of prompting shapes LLM outputs, influencing both cultural preferences and ethical judgments \citep{Vida2024, agarwal2024ethical, whosemorality}. We thus hypothesize that models will be more inconsistent for subjects like ethics, morality, and sociology, which tend to be heavily influenced by sociocultural norms, as opposed to topics with relatively fewer cultural priors, such as mathematics and computer science. The questions in GlobalMMLU are divided into four domains- \textit{STEM, Humanities, Social Sciences} and \textit{Other}. We further subdivide these categories to provide a more detailed analysis. $\kappa_p$ tends to be greater for \textit{STEM} in all the models as opposed to the other subjects (see Figure~\ref{fig:subject_capa}). This affirms our hypothesis about language sensitivity for culturally sensitive domains. Looking at the fine-grained categories (refer Table~\ref{tab:categories}) in Figure ~\ref{fig:capa_scores_gemma_qwen_vertical} we continue to see a substantial difference between $\kappa_p$ of the subjects. 

\subsection{Inter-Model Multilingual Similarity}
\begin{figure*}[t]
    \centering
    \includegraphics[width=0.9\textwidth]{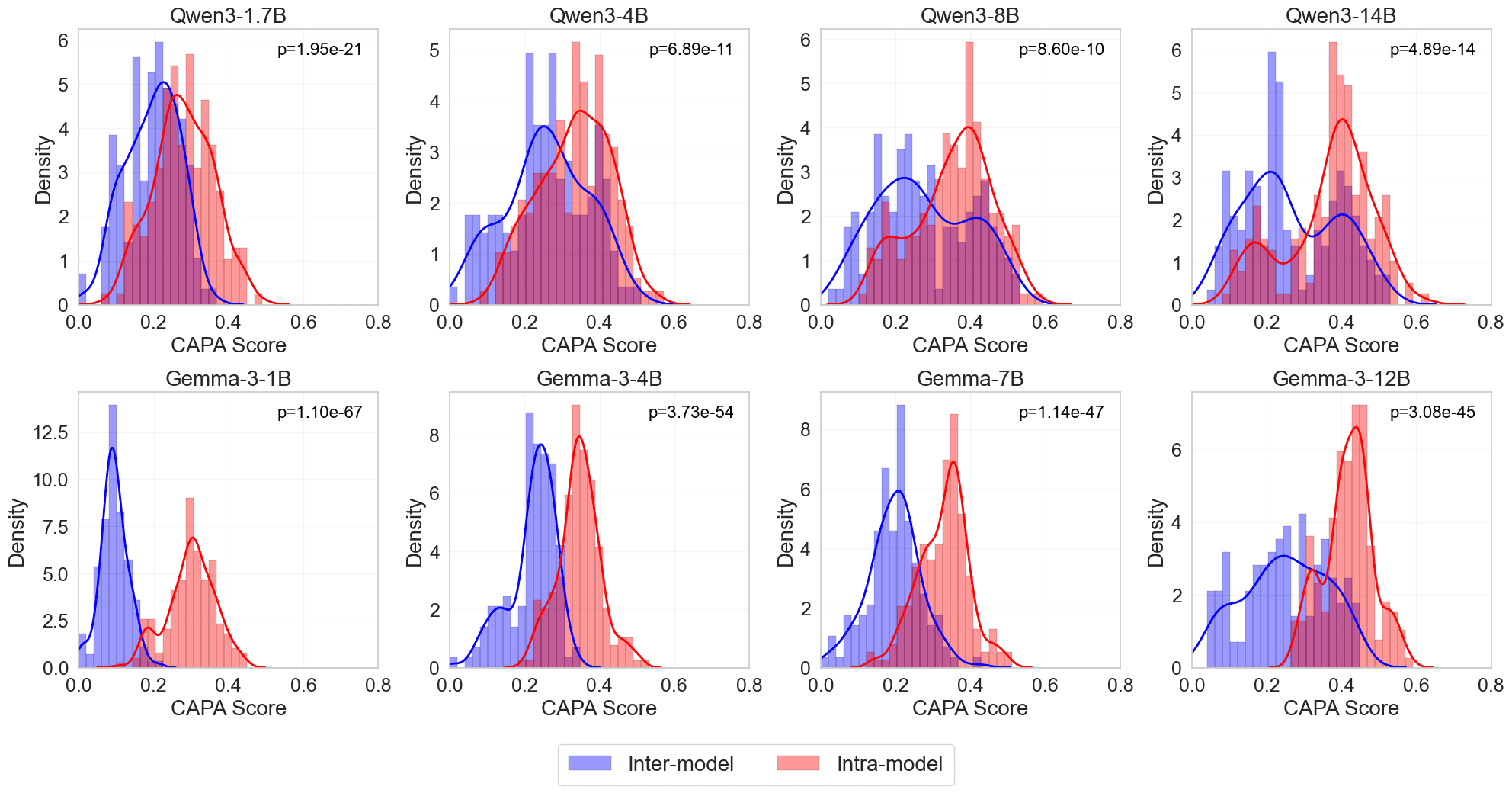}
    \caption{Frequency density distribution of the intra-model (across 20 language pairs) and inter-model (1 model vs remaining 7) $\kappa_p$ scores along with the p-values of the Mann-Whitney U Test. Intra-Model similarity is greater for all models than Inter-Model similarity with high significance.}
    \label{fig:inter_vs_intra}
\end{figure*}

\begin{figure}[H]
    \centering
    \includegraphics[width=\columnwidth]{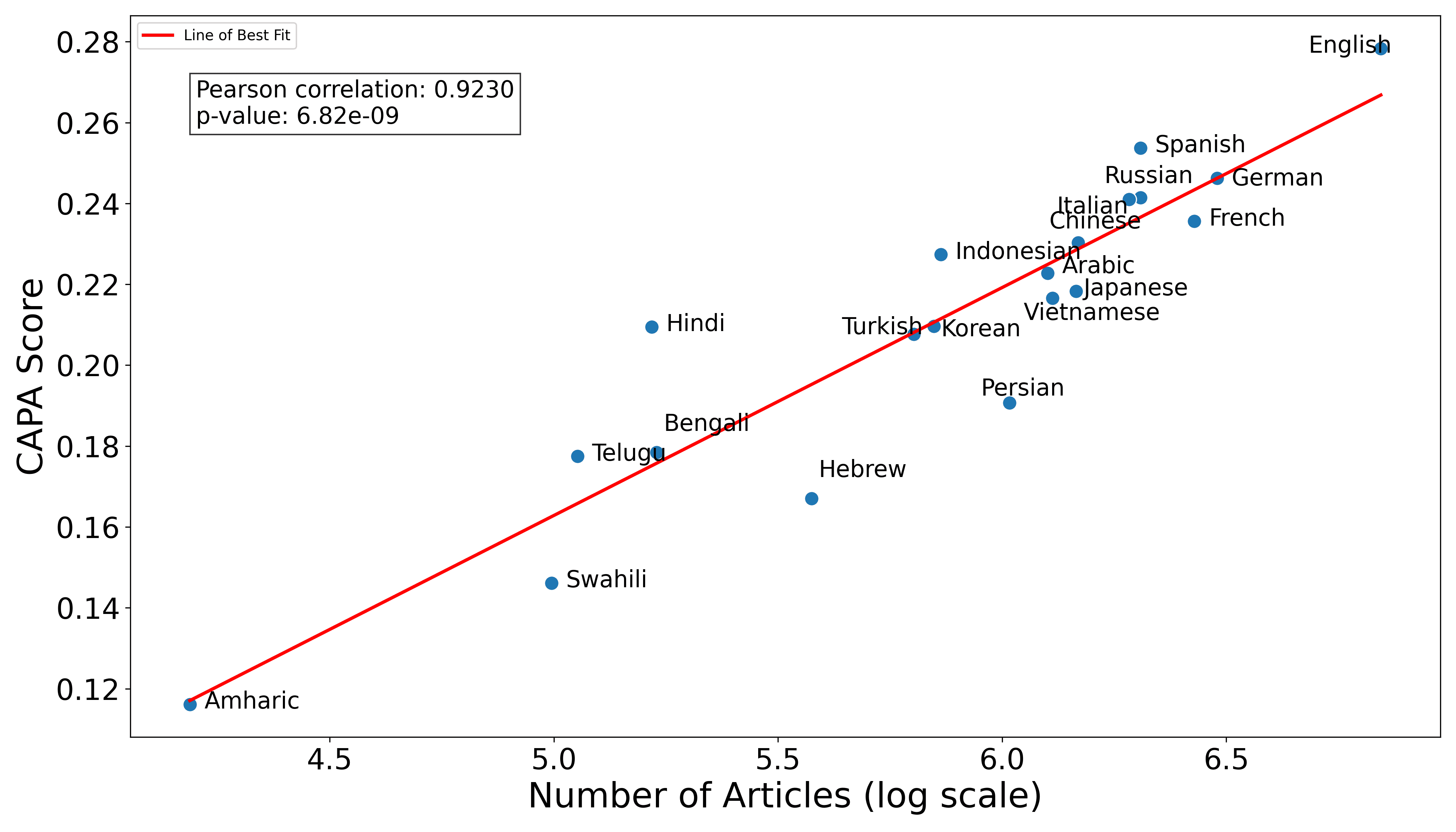}
    \caption{Higher-resource languages exhibit more model agreement. We observe a high correlation between $\kappa_p$ and number of wiki articles (\textit{Pearson correlation} = 0.923).}
    \label{fig:inter-model-CAPA}
\end{figure}

\textbf{RQ3: Do models agree more on high-resource languages?} When we average the $\kappa_p$ scores for a given language across all unique model pairs - a clear trend emerges - high-resource languages tend to have greater inter-model functional similarity, implying that the results are more consistent for languages like English than Amharic across all the models. 
We confirm this by using the number of Wikipedia articles for a given language as a proxy for their resource availability. Figure ~\ref{fig:inter-model-CAPA} indicates a significant positive correlation between the count of Wikipedia articles and inter-model functional similarity $\kappa_p$ score. 

\textbf{RQ4: Is cross-lingual similarity within the same model stronger than cross-model similarity in the same language?} For each model, we find the distribution of the $\kappa_p$ scores for two cases- Intra-Model (across all unique language pairs) and Inter-Model (across all models for each language). For the most part, models tend to be more similar to themselves for different languages than other models for the same language (see Figure ~\ref{fig:inter_vs_intra}). We employ the Mann-Whitney U test \citep{nachar2008mann} - a non-parametric statistical test commonly used to compare two independent samples - for this purpose. The null hypothesis of this test is that randomly selected values from two populations have the same distribution. The p-values (< 0.001) indicate that all tests are statistically significant, confirming that the intra-model and inter-model similarity distributions are significantly different, with intra-model scores tending to be higher. We further conduct an ablation using English, the highest-resource language, as a pivot. The results (see Appendix~\ref{sec:inter_intra_ablation}) remain consistent: intra-model similarity scores are higher than inter-model similarity scores, reinforcing our main findings.
Additionally, we find that the functional and representational similarity correlate to a certain degree in Appendix~\ref{sec:corr_rep_func}.

\section{Conclusion}
We introduced $\kappa_p$ as a functional similarity metric for evaluating multilingual consistency in LLMs. Across GlobalMMLU, we found that larger and more capable models are more consistent across languages, with intra-model similarity exceeding inter-model similarity. Consistency also varies by domain --- being higher in STEM than in culturally sensitive subjects --- and by resource availability, with high-resource languages showing stronger inter-model agreement. Together, these results establish $\kappa_p$ as a practical tool for analyzing multilingual functional behavior beyond accuracy alone. 

\bibliography{main}

\appendix

\section{\texorpdfstring{$\kappa_p$}{kappa	extsubscript{p}} vs Other Metrics}
\label{sec:capa_usage}
We choose the CAPA metric as it has clear advantages over other metrics have been theoretically and empirically validated. $\kappa_p$ metric is chance-adjusted, meaning it is not inflated when model accuracy is high. An example to help understand this is A model with 95\%  accuracy in English and Spanish answers 95/100 questions correctly in both. Raw agreement appears high, but this is trivial—it reflects correctness. $\kappa_p$ downweighs such expected agreement. In contrast, with 50\% accuracy in two low-resource languages, if the model makes similar mistakes, $\kappa_p$ 
 captures this meaningful functional similarity as agreement beyond chance. When we compare it to other metrics Cohen's $\kappa$ and Scott's $\pi$, we observe the difference in inflation due to accuracy. \\

  Consider two raters with predictions [0, 0, 0, 1, 2, 1] and [0, 0, 0, 1, 2, 0] respectively. Let the ground truths be [0, 0, 0, 1, 2, 2]. 

\begin{itemize}
    \item \textbf{Cohen's $\kappa$}
    \[
    \begin{aligned}
    \kappa &= \frac{P_o - P_e}{1 - P_e} = \frac{\tfrac{5}{6} - \tfrac{15}{36}}{1 - \tfrac{15}{36}} \\[6pt]
    &= \frac{0.833 - 0.417}{0.583} \approx 0.714
    \end{aligned}
    \]

    \item \textbf{Scott's $\pi$}
    \[
    \begin{aligned}
    \pi &= \frac{P_o - P_e}{1 - P_e} \\[6pt]
    &\text{where } P_o = \tfrac{5}{6} = 0.833, \\[3pt]
    &P_e = (0.583)^2 + (0.250)^2 + (0.167)^2 \\
    &= 0.431
    \end{aligned}
    \]
    thus
    \[
    \pi = \frac{0.833 - 0.431}{1 - 0.431} \approx 0.707
    \]

    \item \textbf{$\kappa_p$} 
    \[
    \begin{aligned}
    \kappa_p &= \frac{c^{E,M}_{\text{obs}} - c^{E,M}_{\text{exp}}}{1 - c^{E,M}_{\text{exp}}} \\[6pt]
    &\text{where } c^{E,M}_{\text{obs}} = \tfrac{5}{6}, \\[3pt]
    &c^{E,M}_{\text{exp}} = acc_1 \times acc_2 = \tfrac{5}{6} \times \tfrac{5}{6} = \tfrac{25}{36}
    \end{aligned}
    \]
    thus
    \[
    \kappa_p \approx 0.45
    \]
\end{itemize}

\noindent

Since both models are highly accurate (83.3\%), the similarity scores as measured by traditional metrics are inflated. This is not the case with $\kappa_p$ as it takes model accuracy into account. \\

 All the results we have presented for 
 remains consistent for other metrics. These results substantiate our findings, indicating their robustness and generalizability beyond the confines of the $\kappa_p$ metric. These computed values are in tables \ref{tab:pearson_correlation}, \ref{tab:mannwhitney_qwen} and \ref{tab:mannwhitney_gemma}.

 We have also calculated for RankC metric and showed the results here. Consider two raters with probabilistic predictions
\[
R_1 =
\begin{bmatrix}
0.50 & 0.45 & 0.05 \\
0.50 & 0.05 & 0.45 \\
0.05 & 0.45 & 0.50
\end{bmatrix} 
\]

\[
R_2 =
\begin{bmatrix}
0.50 & 0.05 & 0.45 \\
0.50 & 0.45 & 0.05 \\
0.45 & 0.05 & 0.50
\end{bmatrix}.
\]
Finding the maximum probabilities from $R_1$ and $R_2$, the hard labels are
\[
r_1 = [0,0,2], \quad r_2 = [0,0,2].
\]
Thus \(P_o = 1\).

\begin{itemize}
    \item \textbf{Cohen's $\kappa$}
    \[
    \begin{aligned}
    &\text{Rater marginals: } p^{(1)} = p^{(2)} = \left[\tfrac{2}{3},\, 0,\, \tfrac{1}{3}\right], \\[3pt]
    &P_e = \sum_i p^{(1)}_i p^{(2)}_i = \left(\tfrac{2}{3}\right)^2 + 0^2 + \left(\tfrac{1}{3}\right)^2 = \tfrac{5}{9}, \\[6pt]
    &\kappa = \frac{P_o - P_e}{1 - P_e} = \frac{1 - \tfrac{5}{9}}{1 - \tfrac{5}{9}} = 1.0.
    \end{aligned}
    \]

    \item \textbf{Scott's $\pi$}
    \[
    \begin{aligned}
    &\text{Pooled counts over both raters: } [4,\, 0,\, 2] \\
    &p = \left[\tfrac{2}{3},\, 0,\, \tfrac{1}{3}\right], \\[3pt]
    &P_e = \sum_i p_i^2 = \left(\tfrac{2}{3}\right)^2 + 0^2 + \left(\tfrac{1}{3}\right)^2 = \tfrac{5}{9}, \\[6pt]
    &\pi = \frac{P_o - P_e}{1 - P_e} = \frac{1 - \tfrac{5}{9}}{1 - \tfrac{5}{9}} = 1.0.
    \end{aligned}
    \]

    \item \textbf{RankC}
    \[
    \begin{aligned}
    &\text{For each item, let } r^{(1)}, r^{(2)} \text{ be class rankings} \\ &\text{from } R_1, R_2. \text{For} \ j = 1,2,3 \\[6pt]
    &\text{P@}j \;=\;
    \frac{\big|\text{Top-}j(r^{(1)}) \cap \text{Top-}j(r^{(2)})\big|}{j} \\[6pt]
    &\text{Weights: } 
    w_j = \frac{e^{3-j}}{\sum_{\ell=1}^3 e^{3-\ell}} \\
    &\;\;\Rightarrow\;\;
    (w_1,w_2,w_3) \approx (0.665,\,0.245,\,0.090). \\[6pt]
    &\text{From the matrices:} \\ &(\text{P@1},\text{P@2},\text{P@3})=(1,\,0.5,\,1). \\[6pt]
    &\Rightarrow \text{item score} = \sum_{j=1}^3 w_j \cdot \text{P@}j \\[3pt]
    &\quad= 0.665\cdot 1 + 0.245\cdot 0.5 + 0.090\cdot 1 \\[3pt]
    &\quad\approx 0.878.
    \end{aligned}
    \]
    Averaging over all three items (identical here) gives
    \[
    \text{RankC} \approx 0.878.
    \]
    
    \item \textbf{$\kappa_p$} \\
    \[
    \begin{aligned}
    \kappa_p &= \frac{c^{p}_{\text{obs}} - c^{p}_{\text{exp}}}{1 - c^{p}_{\text{exp}}} \\[6pt]
    &\text{where } c^{p}_{\text{obs}} = \; \frac{1}{N} \sum_{i=1}^N \sum_{c=1}^C p^{(1)}_{i,c}\,p^{(2)}_{i,c} = 0.295 \\[3pt]
    &c^{p}_{\text{exp}} = \; \bar p^{(1)}\,\bar p^{(2)} \;+\; \frac{\bigl(1 - \bar p^{(1)}\bigr)\,\bigl(1 - \bar p^{(2)}\bigr)}{C - 1} \\
    &= 0.375
    \end{aligned}
    \]
    thus
    \[
    \text{$\kappa_p$} = -0.128.
    \]

\end{itemize}

\noindent
Collapsing to hard labels yields perfect agreement (\(\kappa=\pi=1.0\)). RankC, which compares top-\(j\) sets from the probability rankings, shows high but non-perfect agreement (\(\approx 0.878\)). $\kappa_p$, which directly evaluates the full probability distributions, detects conflicting uncertainty allocations across classes and therefore yields a \emph{negative} chance-corrected agreement (\(-0.128\)). This is intuitive, as when the models are incorrect, they give very different (and in fact, opposite) predictions which is not captured by the other metrics.

\section{Choice of Languages Used}
\label{sec:lang_choice}
We choose to do our analysis over twenty languages as listed in Table~\ref{tab:lang_table}. The languages chosen belong to a wide range of groups, including the Afroasiatic (Amharic, Arabic, Hebrew), Dravidian (Telugu), Germanic (English, German), and Indo-Iranian (Persian, Hindi, Bengali) language families/branches, among others. The subset of GlobalMMLU was curated to represent a spectrum of resource availability, where high-resource languages refer to those with abundant linguistic data, such as large corpora, annotated datasets, and digital tools (e.g., English, Spanish), while low-resource languages lack such resources and infrastructure (e.g., Amharic, Telugu). This selection allows us to assess model behaviour across typologically and resource-diverse settings. All the languages have an equal number of questions, and we have chosen the subset among these which have consistent answers among all the languages leading to a total of 13844 questions in each language. 

\section{Sub-Categorization of GlobalMMLU}

We sub-categorized the existing categories of GlobalMMLU to make better and fine-grained inferences. We follow the standard GlobalMMLU setup in lm-eval-harness \citep{eval-harness} to conduct the evaluations. The tables \ref{tab:super_categories} and \ref{tab:categories} show the categorisation based on the four domains and further split 14 categories, respectively. The tables also show the distribution of the samples for each category. Each numerical value in the Samples columns of the table corresponds to the number of resulting samples for a given model for a given language.

\section{Ablations for Inter-model vs Intra-model Similarity}
\label{sec:inter_intra_ablation}
\begin{figure*}[h]
    \centering
    \begin{subfigure}{\textwidth}
        \centering
        \includegraphics[width=\textwidth]{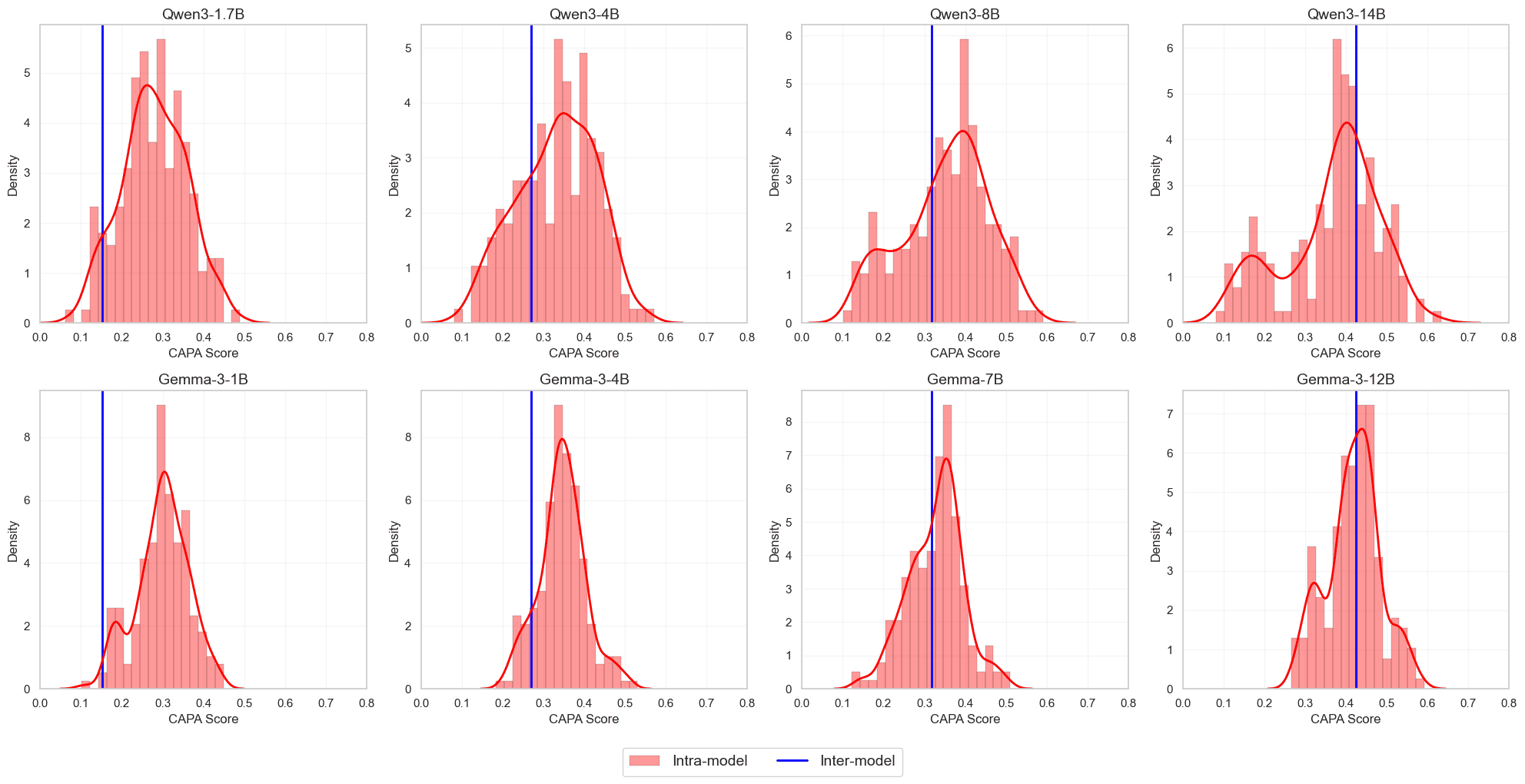}
        \caption{Frequency distribution of the intra-model (across 20 language pairs) and inter-model (1 model vs closest family model).}
        \label{fig:inter_vs_intra_ablation_all}
    \end{subfigure}

    \vspace{1em}

    \begin{subfigure}{\textwidth}
        \centering
        \includegraphics[width=\textwidth]{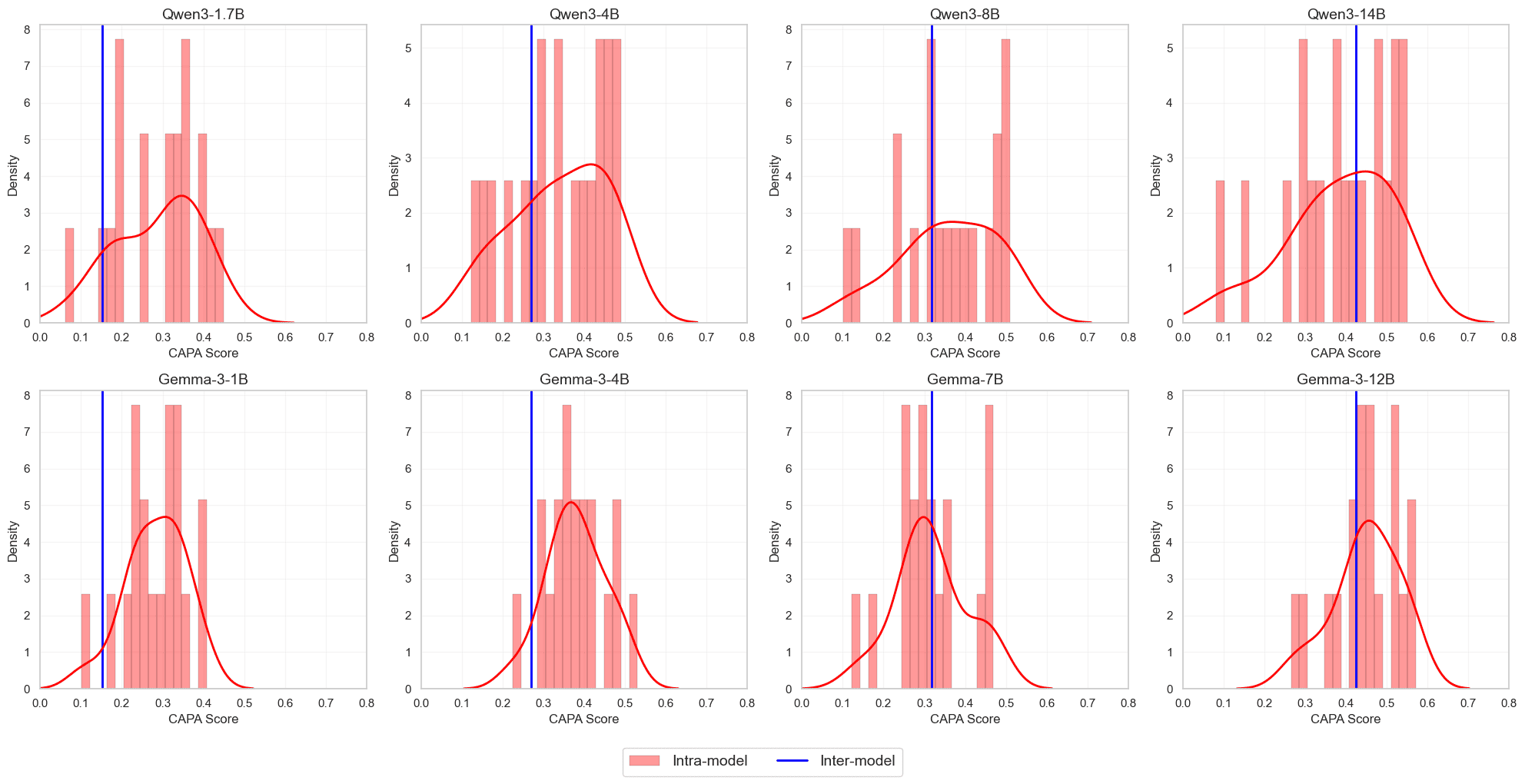}
        \caption{Frequency distribution of the intra-model (across English-non-English pairs) and inter-model (1 model vs closest family model).}
        \label{fig:inter_vs_intra_ablation_en}
    \end{subfigure}
\end{figure*}

We explore an alternate way to plot inter-model similarity by removing potential confounders from cross-size comparisons. Initially, the computation for the inter-model similarity was plotting the distribution of the computed $\kappa_p$ values for each model with the remaining seven models across 20 languages. For intra-model similarity, we compute, for each model, the distribution of $\kappa_p$ values across 20 unique language pairs. For this ablation, we compute the $\kappa_p$ values for inter-model similarity to be a single $\kappa_p$ value for each model with the model of the other family with the closest number of parameters (model size). We then plot two distributions for intra-model similarity. In Figure \ref{fig:inter_vs_intra_ablation_all}, the intra-model similarity computation remains the same, calculation $\kappa_p$ across 20 unique language pairs. In Figure \ref{fig:inter_vs_intra_ablation_en}, the intramodel similarity distribution has been revised to include only pairs of English-non-English languages. ($en-\{lang\}$). The results remain consistent with previous results, showing that intra-model similarity is still greater than inter-model similarity.

\section{Some Correlation Between Functional and Representational Similarity}
\label{sec:corr_rep_func}
Following the procedure in \citet{wu2024semantic}, we compute the representation cosine similarity and use the last token position as the sentence representation over a subset of the translation dataset, FLORES-101 \citep{goyal2022flores}. We subtract these scores by a baseline of non-matching sentences and find that when two languages have a greater $\kappa_p$ score, i.e. they have high functional similarity, they also tend to have a greater representational similarity as measured by the increase over the baseline. We do it over limited layers of the Qwen model (Qwen3-4B and Qwen3-8B) due to compute constraints. This experiment is carried out to establish some degree of correlation between the two notions of similarity, the existence of which has been debated before in \citet{klabunde2025similarity}.

\begin{figure}[h]
    \centering
    \includegraphics[width=\columnwidth]{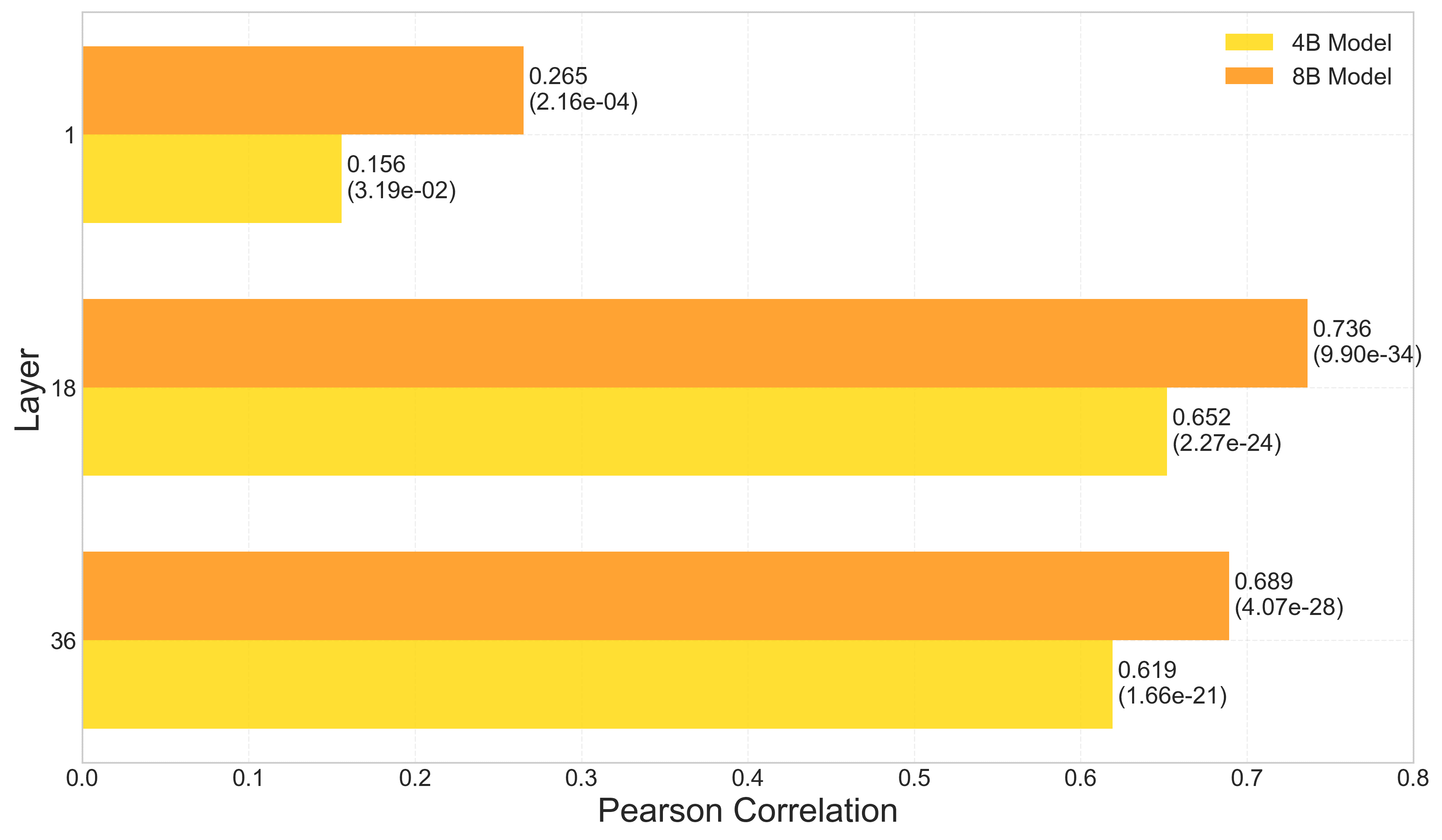}
    \caption{Languages with higher functional similarity ($\kappa_p$) also exhibit greater representational similarity. Representation cosine similarity is computed using the last token position from FLORES-101 sentence pairs. Scores are baseline-adjusted using non-matching sentence pairs.}
    \label{fig:correlation_layer}
\end{figure}

\section{Limitations}
Although our findings establish statistically significant correlations across languages and models, we cannot establish causality for the observed phenomena as this would require extensive mechanistic interventions. $\kappa_p$ is limited to multiple-choice benchmarks, and there is a lack of free-form functional similarity metrics that take error consistency into account. This restricts our study to multilingual MCQ benchmarks. However, there is also a lack of parallel multilingual MCQ benchmarks, and most existing ones, such as \citep{mmluprox}, are variants of MMLU. Hence, we limit our analysis to the largest of these, GlobalMMLU.

\section{Future Work}
We advocate for $\kappa_p$ to be used as a tool for analyzing multilinguality. We find interesting observations on the GlobalMMLU dataset, and feel that using this approach would be beneficial to the field of multilingual NLP in addition to the substantial work already being done in the representational space. There is also a great scope to explore if the two notions of similarity have any fundamental connection. 

Although we hypothesize that having more data could help in improving multilingual consistency, it is also possible that it is inherently easier to learn one language from a greater capacity in another language if their underlying structures are similar. Is the cause of high functional similarity between two languages a function of their training (multilingual or parallel corpus), a natural alignment or a common syntactic structure of the two languages, or something different altogether? Establishing causality to our observations using interpretability techniques would be challenging but worthwhile. 

Besides our current use case, we can see it being valuable in several applications. Higher functional similarity between two languages can have consequences on downstream tasks. For example, if a model with high $\kappa_p$ between Hindi and English exists, it might become easier to translate between the two languages. Furthermore, it might allow such models to interpret Hindi-English code mixed text samples more easily than another pair with a lower score.

\label{sec:lang_table}
\begin{table*}[h]
  \centering
  \vspace{0.25em}
  \begin{tabular}{cl}
    \hline
    \textbf{Code} & \textbf{Language} \\
    \hline
    am & Amharic \\
    ar & Arabic \\
    bn & Bengali \\
    zh & Chinese \\
    en & English \\
    \hline
  \end{tabular}
  \hspace{0.25em}
  \begin{tabular}{cl}
    \hline
    \textbf{Code} & \textbf{Language} \\
    \hline
    fr & French \\
    de & German \\
    he & Hebrew \\
    hi & Hindi \\
    id & Indonesian \\
    \hline
  \end{tabular}
  \hspace{0.25em}
  \begin{tabular}{cl}
    \hline
    \textbf{Code} & \textbf{Language} \\
    \hline
    it & Italian \\
    ja & Japanese \\
    ko & Korean \\
    fa & Persian \\
    ru & Russian \\
    \hline
  \end{tabular}
  \hspace{0.25em}
  \begin{tabular}{cl}
    \hline
    \textbf{Code} & \textbf{Language} \\
    \hline
    es & Spanish \\
    sw & Swahili \\
    te & Telugu \\
    tr & Turkish \\
    vi & Vietnamese \\
    \hline
  \end{tabular}
  \hspace{0.25em}
  \caption{Language codes and their corresponding language names used in our experiments.}
  \label{tab:lang_table}
\end{table*}

\begin{table*}[h]
\centering
\renewcommand{\arraystretch}{1.3}
\small
\begin{tabular}{lccc}
\toprule
\textbf{Metric / Evaluation} &
\textbf{Pearson correlation for Size} &
\textbf{Pearson correlation for Accuracy} &
\textbf{\valp{Pearson correlation for Resource}{log no. of Articles}} \\
\midrule
$\kappa_p$         & \valp{0.7864}{0.02062} & \valp{0.7884}{0.02009} & \valp{0.9230}{6.82e-09} \\
Cohen's $\kappa$   & \valp{0.8862}{0.003376} & \valp{0.9714}{5.694e-05} & \valp{0.9321}{2.28e-09} \\
Scott's $\pi$      & \valp{0.8861}{0.003385} & \valp{0.9714}{5.728e-05} & \valp{0.9313}{2.53e-09} \\
\bottomrule
\end{tabular}
\caption{Pearson correlation coefficients (top) with p-values in parentheses (bottom).}
\label{tab:pearson_correlation}
\end{table*}

\begin{table*}[h]
\centering
\renewcommand{\arraystretch}{1.3}
\small
\begin{tabularx}{\textwidth}{lXXXX}
\toprule
\textbf{Metric} & \textbf{Qwen3-1.7B} & \textbf{Qwen3-4B} & \textbf{Qwen3-8B} & \textbf{Qwen3-14B} \\
\midrule
$\kappa_p$       & \valp{12150}{1.95e-21} & \valp{16686}{6.89e-11} & \valp{17278}{8.60e-10} & \valp{15148}{4.89e-14} \\
Cohen's $\kappa$ & \valp{23750}{6.08e-02} & \valp{21710}{1.29e-03} & \valp{17992}{1.48e-08} & \valp{14766}{6.90e-15} \\
Scott's $\pi$    & \valp{22358}{5.26e-03} & \valp{21644}{1.11e-03} & \valp{17542}{2.53e-09} & \valp{14460}{1.38e-15} \\
\bottomrule
\end{tabularx}
\caption{Mann–Whitney U statistics for Qwen models (p-values in parentheses).}
\label{tab:mannwhitney_qwen}
\end{table*}

\begin{table*}[h]
\centering
\renewcommand{\arraystretch}{1.3}
\small
\begin{tabularx}{\textwidth}{lXXXX}
\toprule
\textbf{Metric} & \textbf{gemma-3-1b-it} & \textbf{gemma-3-4b-it} & \textbf{gemma-7b} & \textbf{gemma-3-12b-it} \\
\midrule
$\kappa_p$       & \valp{180}{1.10e-67}   & \valp{3050}{3.73e-54}  & \valp{4556}{1.14e-47}  & \valp{5148}{3.08e-45} \\
Cohen's $\kappa$ & \valp{5660}{3.46e-43}  & \valp{11650}{7.83e-23} & \valp{16394}{1.88e-11} & \valp{7316}{6.87e-37} \\
Scott's $\pi$    & \valp{4364}{1.79e-48}  & \valp{11176}{3.37e-24} & \valp{15598}{4.53e-13} & \valp{7228}{3.27e-37} \\
\bottomrule
\end{tabularx}
\caption{Mann–Whitney U statistics for Gemma models (p-values in parentheses).}
\label{tab:mannwhitney_gemma}
\end{table*}

\begin{table*}[h]
\centering
\renewcommand{\arraystretch}{1.3}
\small
\begin{tabular}{lp{11cm}r}
\toprule
\textbf{Domain} & \textbf{Subjects} & \textbf{\# Samples} \\
\midrule
STEM & College Chemistry, High School Computer Science, College Biology, Abstract Algebra, High School Mathematics,
      Computer Security, Machine Learning, College Physics, Conceptual Physics, Astronomy,
      High School Biology, High School Physics, Anatomy, College Mathematics, Electrical Engineering,
      College Computer Science, High School Chemistry, High School Statistics, Elementary Mathematics & 3153 \\
\cmidrule(lr){1-3}
Humanities & Philosophy, World Religions, Professional Law, Moral Scenarios, High School European History,
           Moral Disputes, Jurisprudence, Formal Logic, High School US History, Prehistory,
           High School World History, International Law, Logical Fallacies & 4511 \\
\cmidrule(lr){1-3}
Social Sciences & High School Microeconomics, High School Geography, US Foreign Policy, Professional Psychology,
                 Security Studies, High School Government and Politics, High School Psychology, Econometrics,
                 Sociology, High School Macroeconomics, Public Relations, Human Sexuality & 3076 \\
\cmidrule(lr){1-3}
Other & Professional Accounting, Professional Medicine, College Medicine, Marketing, Nutrition,
       Global Facts, Clinical Knowledge, Human Aging, Virology, Miscellaneous,
       Business Ethics, Management, Medical Genetics & 3104 \\
\bottomrule
\end{tabular}
\caption{Original Grouping of GlobalMMLU subjects into 4 domains with corresponding sample counts.}
\label{tab:super_categories}
\end{table*}

\begin{table*}[h]
\centering
\renewcommand{\arraystretch}{1.3}
\small
\begin{tabular}{lp{9cm}r}
\toprule
\textbf{Category} & \textbf{Subjects} & \textbf{\# Samples} \\
\midrule
Mathematics & Abstract Algebra, College Mathematics, Elementary Mathematics \newline
            High School Mathematics, High School Statistics, Formal Logic \newline
            Logical Fallacies & 1064 \\
\cmidrule(lr){1-3}
Logic     & Formal Logic,
            Logical Fallacies & 289 \\
\cmidrule(lr){1-3}
Physics     & College Physics, Conceptual Physics, High School Physics, Astronomy & 640 \\
\cmidrule(lr){1-3}
Biology     & College Biology, High School Biology, Human Aging \newline
            Human Sexuality, Virology & 971 \\
\cmidrule(lr){1-3}
Chemistry   & College Chemistry, High School Chemistry & 303 \\
\cmidrule(lr){1-3}
Medicine    & Anatomy, Clinical Knowledge, College Medicine \newline
            Medical Genetics, Nutrition, Professional Medicine & 1251 \\
\cmidrule(lr){1-3}
Computer Science & College Computer Science, High School Computer Science \newline
                 Computer Security, Machine Learning & 412 \\
\cmidrule(lr){1-3}
Economics and Business & Econometrics, High School Macroeconomics, High School Microeconomics \newline
                       Business Ethics, Management, Marketing \newline
                       Professional Accounting & 1461 \\
\cmidrule(lr){1-3}
Psychology and Sociology & High School Psychology, Professional Psychology, Sociology & 1358 \\
\cmidrule(lr){1-3}
Geography and Global Affairs & Global Facts, High School Geography, US Foreign Policy \newline
                             Security Studies & 643 \\
\cmidrule(lr){1-3}
History     & High School US History, High School European History \newline
            High School World History, Prehistory & 741 \\
\cmidrule(lr){1-3}
Government and Law & High School Government and Politics, International Law, Jurisprudence \newline
                   Professional Law & 1951 \\
\cmidrule(lr){1-3}
Philosophy and Ethics & Philosophy, Moral Disputes, Moral Scenarios & 1552 \\
\cmidrule(lr){1-3}
Miscellaneous & World Religions, Public Relations, Electrical Engineering, Miscellaneous & 1208 \\
\bottomrule
\end{tabular}
\caption{Fine-grained categorization of GlobalMMLU subjects used in our ablation.}
\label{tab:categories}
\end{table*}

\end{document}